\documentclass[sigconf]{acmart}

\AtBeginDocument{%
  }

\setcopyright{cc}
\setcctype{by}
\copyrightyear{2026}
\acmYear{2026}
\acmDOI{10.1145/3829333.3847960}

\acmConference[SA Posters '26]{SIGGRAPH Asia 2026 Posters}{December 01--04, 2026}{Kuala Lumpur, Malaysia}
\acmISBN{979-8-4007-2839-6/2026/12}

\usepackage{cuted}
\usepackage{capt-of}

\usepackage{tikz}
\usetikzlibrary{calc}

\begin{document}

\title{CreativeFlow: A One-to-Many Analogical Relation Transfer Method for 3D Asset Generation}


\author{Xuechen Li}
\affiliation{
  \institution{College of Design and Innovation, Tongji University}
  \city{Shanghai}
  \country{China}
}
\email{primav1022@tongji.edu.cn}

\author{Shuai Zhang}
\affiliation{
  \institution{College of Design and Innovation, Tongji University}
  \city{Shanghai}
  \country{China}
}
\email{zhangshuai2@tongji.edu.cn}

\author{Nanxuan Zhao}
\affiliation{
  \institution{Adobe Research}
  \city{San Jose, California}
  \country{United States}
}
\email{nanxuanzhao@gmail.com}

\author{Qing Chen}
\authornote{Corresponding author.}
\affiliation{
  \institution{College of Design and Innovation, Tongji University}
  \city{Shanghai}
  \country{China}
}
\email{jane.qing.chen@gmail.com}

\renewcommand{\shortauthors}{Li et al.}

\begin{abstract}
    Inspired by cognitive science, we present \textsc{CreativeFlow}, an analogical generation framework that explicitly models analogical divergent thinking to mitigate creative homogenization in text-to-3D pipelines. Our method derives a series of meaningful yet relationally similar source-target asset pairs, each featuring distinct geometric configurations. Expert evaluations demonstrate that our framework substantially enhances creative novelty and visual fascination. This workflow and its resulting assets establish a foundational dataset and benchmark for future relation-aware 3D model training.
  
\end{abstract}

\vspace{-5pt}


\vspace{-20pt}
\vspace{-20pt}


\vspace{-5pt}
\begin{teaserfigure}
 \includegraphics[width=\textwidth]{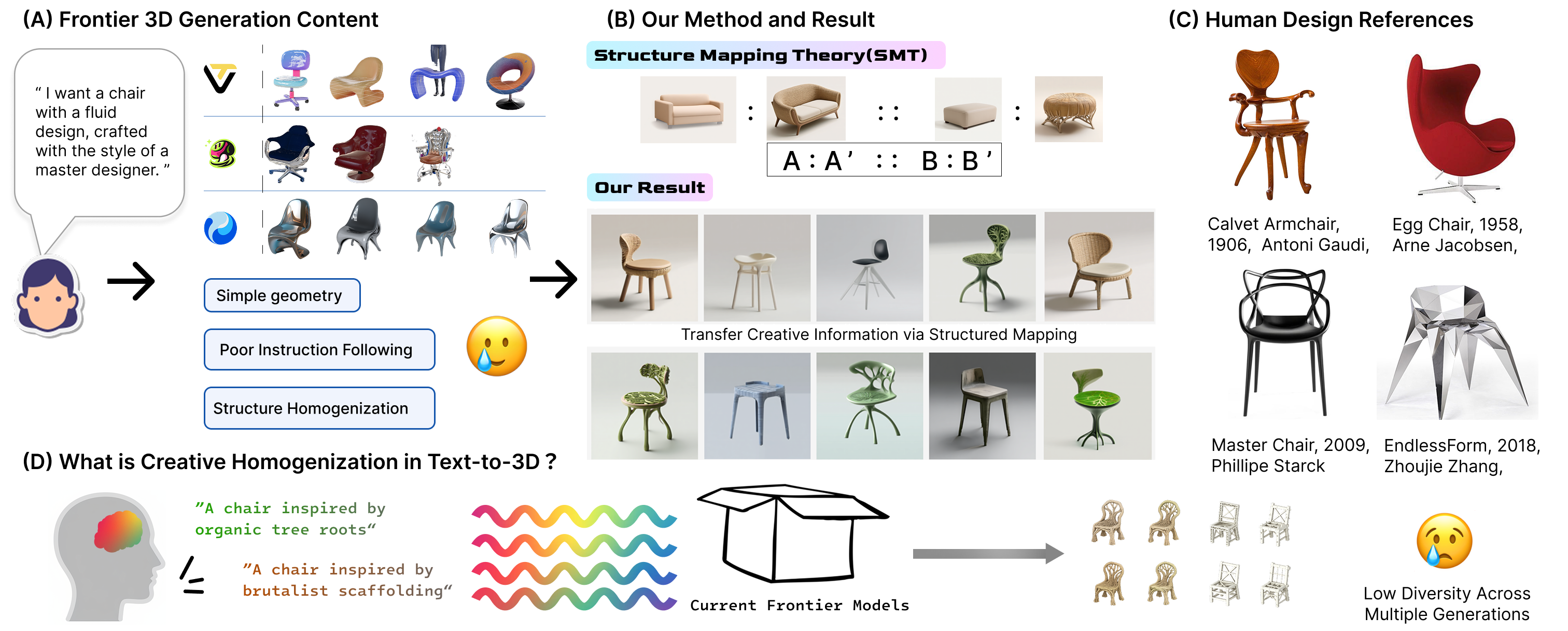}
  \caption{(A) Standard text-to-3D models that collapse diverse intents into homogenized outputs, (B) \textsc{CreativeFlow} generates structurally diverse 3D designs approaching the richness of human references (C) and the illustration of creative homogenization(D).}
  \Description{(A) Standard text-to-3D models that collapse diverse intents into homogenized outputs, (B) CreativeFlow generates structurally diverse 3D designs approaching the richness of human references (C) and the illustration of creative homogenization(D).}
  \label{fig:teaser}
\end{teaserfigure}

\maketitle

\vspace{-5pt}
\section{Introduction}

Current text-to-3D systems generate high-fidelity assets but struggle with open-ended, creative prompts, often collapsing outputs into semantically conservative, homogenized forms (Fig.~\ref{fig:teaser}-(A)). This creative homogenization stems from limited semantic transfer; current pipelines typically produce shallow style or color variations rather than exploring novel structural deviations of the source object~\cite{li2026beyond}.

We address this bottleneck through analogical creativity~\cite{holyoak1996mental}, a classical mechanism in cognitive science~\cite{lu2019emergence}. Grounded in \textbf{Structure Mapping Theory (SMT)}~\cite{gentner1983structure}, we formulate creative prompts as relational mapping problems:\textbf{ $A:A'::B:B'$} (Fig.~\ref{fig:teaser}-(B)), where the source object $A$ maps its attributes or structures $A'$ to a target design $B'$ via an analogical concept $B$. This formulation guides generative pipelines to transfer meaningful structural, functional, or material logic rather than superficial, surface-level details.

We introduce \textsc{CreativeFlow}, a one-to-many analogical generation framework that maps a single source object to diverse, relation-aware target designs. Our method combines structured relational retrieval from external knowledge graphs (spanning biomimetic structures and art styles) with the inherent associations of large language models. After mapping inference, candidates are filtered via a multi-objective scoring engine evaluating semantic novelty, 3D feasibility, and structural transferability, and then translated into explicit instructions for 2D-prior-to-3D generation. This poster presents this expansion pipeline along with preliminary user studies validating its effectiveness in enhancing perceived design diversity and creative structural alignment.
\begin{figure}
    \centering
    \includegraphics[width=1\linewidth]{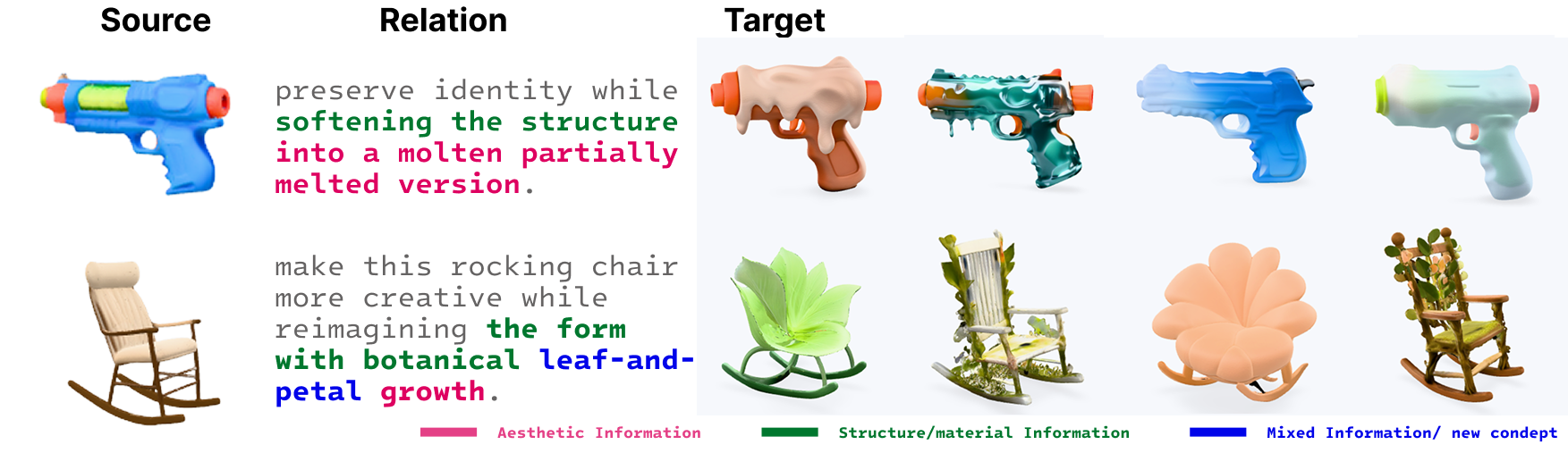}
    \caption{Examples of relation-guided analogical transfer. \textsc{CreativeFlow} transfers explicit aesthetic, structural/material, and mixed relations from a source asset to generate diverse yet recognizable target designs.}
    \label{fig:result1}
\end{figure}

\begin{figure*}
    \centering
    \includegraphics[width=1\linewidth]{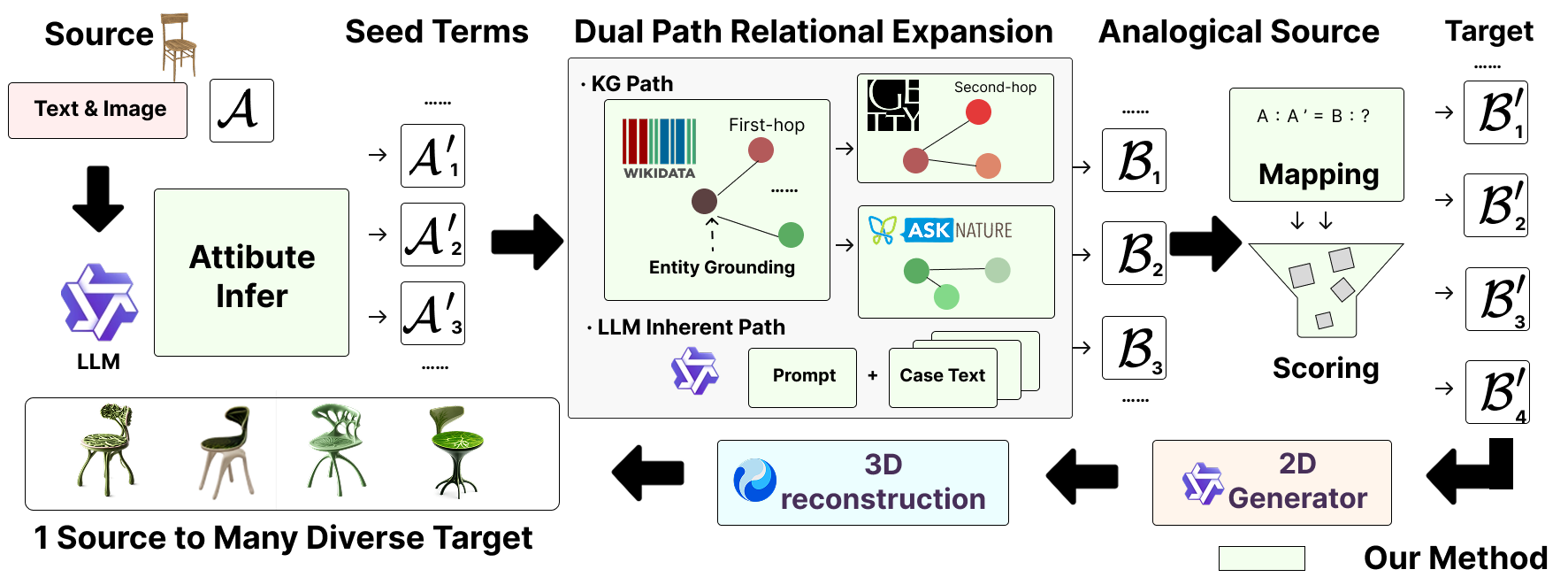}
    \caption{Our pipeline modeling $A : A' :: B : B'$. We extract source attribute set ${A'}$ from $A$ via Attribute Inference, retrieve analogical concept set $B$ via Dual-Path Expansion, and solve for target attribute set $B'$ to reconstruct diverse 3D target assets.}
    \label{method}
    \vspace{-5pt}
\end{figure*}

\vspace{-5pt}
\section{Methodology}

As shown in Fig.~\ref{fig:teaser}, when faced with ambiguous semantic inputs with multiple interpretations, existing 3D generative models tend to converge toward dominant semantic modes, concentrating different sampled results in similar local solution spaces and failing to explore alternative design directions guided by other semantic relations. This limitation arises because current pipelines do not explicitly model transferable relations to map local components to cross domain concepts. To address this, we propose a relation driven, cognitive inspired analogical expansion framework.

\textsc{CreativeFlow} translates a single source asset $\mathcal{A}$ into a diverse target manifold through three stages (Fig.~\ref{method}): attribute inference, dual-path relational expansion, and analogical mapping with 3D asset generation. First, given a source $A$ (e.g., a chair), an LLM performs attribute inference across dimensions of function, material, structure, visuals, and affordances~\cite{qian1996function}. This stage parses the geometry and functional regions of the source to extract core transferable parts as seed terms $A'$ (e.g., support frame, seat shell, backrest, or surface material), anchoring subsequent cross-domain relational mapping.

To identify analogical sources $B$ and their transferable characteristics, the dual-path expansion module routes seed terms $A'$ by category into two paths. The first path, KG expansion, grounds both the source asset and attributes in \textit{Wikidata}~\cite{vrandevcic2014wikidata} to locate first-hop neighborhood nodes, then executes second-hop queries along \textit{Getty AAT}~\cite{harpring2010development} and \textit{AskNature}~\cite{deldin2013asknature} to retrieve hyponyms. Specifically, \textit{Getty AAT} expands $\mathcal{B}$ across style, craft, material, ornament, and object type to provide aesthetic attributes. Meanwhile, \textit{AskNature} concentrates on biological structures and biomimetic mechanisms, expanding $B$ across support, protection, attachment, lightweight structure, and surface function to yield physically groun-ded natural strategies. The second path is an LLM-internal knowledge transfer method based on prompt expansion of historical transfer cases.

Finally, we model creative transfer as solving a proportional analogy equation grounded in Structure Mapping Theory (SMT): $A : A' = B : B'$, where $B'$ is the transferable characteristic attribute of the analogical source $B$ mapping to $A'$. For example, mapping $\text{chair} : \text{support frame} = \text{coral} : \text{branching structure}$ transfers the coral's branching geometry to the chair's structural support. To ensure target diversity and physical viability, a multi-objective engine filters candidate mappings ($B'$) by balancing semantic novelty, 3D feasibility, and structural transferability. These selected targets are then converted into explicit part-level instructions (e.g., ``coral-branch support frame''), processed by a 2D Generator to synthesize view-consistent visual priors, and finally reconstructed by a 3D Reconstructor into 3D meshes, achieving end-to-end generation from a single prototype to diverse assets.

\vspace{-5pt}
\section{Preliminary Evaluation and Results}

As shown in Fig.~\ref{method}, \textsc{CreativeFlow} expands a single source asset (comprising text, image, and mesh) into 5 to 10 relations with aesthetic and structural information, each yielding targets with diverse geometric deformations, theoretically enabling infinite creative designs (As illustarted in Fig.\ref{fig:result1}).

For evaluation, we generated 3D assets from matched prompts and compared \textsc{CreativeFlow} with baseline generation pipelines through two expert user studies using five-point scales against three baselines (\textit{Hunyuan3D-2, TRELLIS-2}, and \textit{SF3D}). In the identity-preservation and novelty study, our method achieved mean scores of $3.66$ and $3.86$, respectively, demonstrating its ability to produce structurally novel yet recognizable assets. In the controlled visual-fascination study, ten design experts rated our method at $3.31$ out of $5$, compared with $2.84$ for the strongest baseline, supporting its greater visual and conceptual appeal.

\begin{figure*}
    \centering
    \includegraphics[width=1\linewidth]{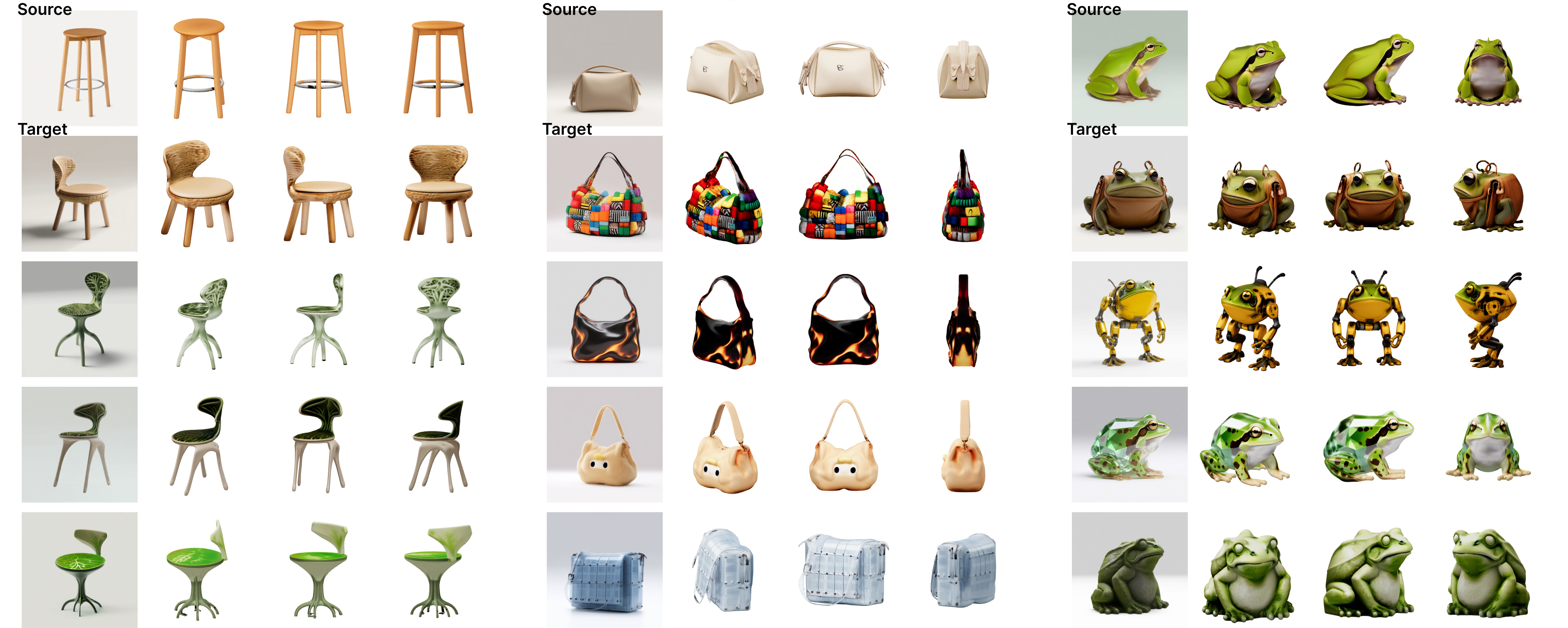}
    \caption{\textbf{More results.} One-to-many qualitative results across stools, handbags, and frogs. For each source, different analogical relations produce diverse target designs, with each row showing the source and multi-view renderings of the generated target.}
    \label{fig:result2}
\end{figure*}

\vspace{-5pt}
\section{Future Work}

This poster represents an initial step toward relation driven 3D creative generation~\cite{wang2024multi}. In the Future, First, we aim to establish robust evaluation metrics specifically designed to quantify generative creativity and analogical structural rationality, filling the gap in standardized creativity evaluation. Second, we will construct a large-scale, structured source-relation-target dataset dedicated to training and benchmarking future model~\cite{deitke2023objaverse}. Finally, we plan to develop next-generation human-computer interactive creative tools that integrate our analogical expansion pipeline into collaborative design workflows, enabling intuitive, controllable, and inspiring human-AI ideation.

\vspace{-10pt}
\bibliographystyle{ACM-Reference-Format}
\bibliography{sample-base}

\end{document}